# Applying Deep Neural Networks to automate visual verification of manual bracket installations in aerospace

John Oyekan, Liam Quantrill, Christopher Turner, Ashutosh Tiwari

***Abstract*—** In this work, we explore a deep learning based automated visual inspection and verification algorithm, based on the Siamese Neural Network architecture. Consideration is also given to how the input pairs of images can affect the performance of the Siamese Neural Network. The Siamese Neural Network was explored alongside Convolutional Neural Networks. In addition to investigating these model architectures, additional methods are explored including transfer learning and ensemble methods, with the aim of improving model performance. We develop a novel voting scheme specific to the Siamese Neural Network which sees a single model vote on multiple reference images. This differs from the typical ensemble approach of multiple models voting on the same data sample. The results obtained show great potential for the use of the Siamese Neural Network for automated visual inspection and verification tasks when there is a scarcity of training data available. The additional methods applied, including the novel similarity voting, are also seen to significantly improve the performance of the model. The metrics evaluated indicate that the Siamese Neural Network's performance nears a level that would be acceptable in industry. We apply the publicly available omniglot dataset to validate our approach. According to our knowledge, this is the first time a detailed study of this sort has been carried out in the automatic verification of installed brackets in the aerospace sector via Deep Neural Networks.

Dataset: https://github.com/oyekanjohn/Brackets_data

## 1.0 Introduction

The demand on the aviation industry to provide more efficient aircraft, in higher volumes, while maintaining safety and high standards is challenging even for world-leading aircraft manufacturers [1]. Majority of the manufacturing work in the aerospace industry is still conducted manually. For example, the manual installation of brackets that secure electrical and hydraulic lines is one of such tasks that take place in enclosed places of an airplane' wing. This requires a worker to glue a bracket to the wing securely, within a marked boundary and with only limited visibility (See Figure 1 and 2). The conditions for a correct installation of a bracket are such that the bracket must be within the boundaries of the required area, within a tolerance, and that there must be flow of adhesive around the whole circumference of the bracket (See Figure 2). After the gluing process, the conditions for a correct installation are visually checked to ensure that they are met. Since the aircraft assembly stage is still manually intensive [2], it is difficult to upscale the production of aircraft to meet high demand without overstretching the workforce and jeopardizing quality and safety. Therefore, many aircraft manufacturers, are investigating ways to digitize the manual assembly stage of aircraft production [1]. This is particularly true in the areas of the installation of components into the aircraft wings, where work is conducted in confined spaces as seen in Figure 1 [3]. There have been some studies on the installation of components into aircraft wings, however there is little research into the verification of such installations. Automated verification by visual inspection is commonplace in high-volume, low-variation manufacturing using methods like machine vision and deep learning. Nevertheless, this technology is yet to be widely adopted by high value manufacturing industries, like the aircraft manufacturing industry, especially in scenarios shown in Figure 1. Since manual visual verification of installed brackets in enclosed places like Figure 1 is challenging, we aim to develop and compare two deep learning architectures, a Convolutional Neural Network and a Siamese Neural Network, for the purposes of automated visual verification of manually installed bracket components. According to our current knowledge, this is the first time a detailed study of this sort has been carried out in the automatic verification of installed brackets in the aerospace sector via Deep Neural Networks.

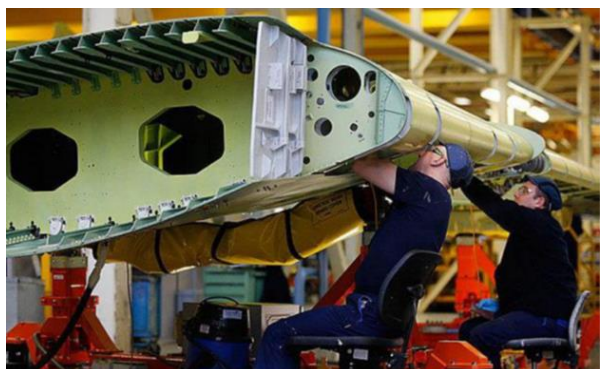
**Figure 1.** Component installations in an aircraft wing.

In Section 2, a literature review is presented to assess the current state of visual inspection in the aircraft manufacturing industry, and to identify suitable methods to tackle the problem of automated verification of manually installed brackets. Section 3 presents the methodology, including the dataset used, the actual network architectures, training strategies, and the specific tests conducted. Section 4 presents the experiments conducted as well as a critical discussion of the results. The implications of the results in an industrial context are discussed in Section 5 while in Section 6, conclusions are drawn on both the academic and industrial implications of the work; limitations of the research and future work are also highlighted.

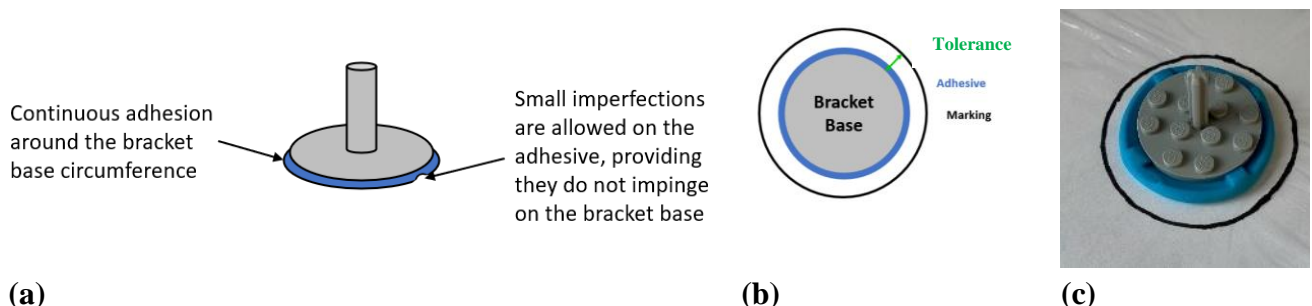

(a)     (b)     (c)

**Figure 2.** Adhesive bracket application criteria in the Aerospace industry (a), with top view tolerance (b) and replica bracket setup used for experimental data generation (c).

## 2.0   Literature review

In this section, the current state of visual inspection within manufacturing is discussed and the state-of-the-art methods identified. Current research into visual inspection in the aircraft manufacturing industry is also evaluated to identify research gaps. These gaps are then used to inform further research into specific methods, particularly deep learning methods, for implementation and testing in this work.

### 2.1 Industrial Visual Inspection Techniques

Visual inspection of manufactured products to ensure they meet standards have been a norm for a very long time. Visual inspection was originally, and in many cases is still carried out manually by human workers across a range of industries, even though it has been shown that humans are prone to making many errors due to various social and environmental factors [4][5]. The manufacturing industry has therefore started to adopt automated approaches to visual inspection to overcome these shortfalls of human workers. However the primary applications of this are in high-volume, low-variance mass manufacturing [6]. Despite human flaws, manual visual inspection is still carried out in low-volume, high-value manufacturing such as aircraft wing production, where the consequences of inspection errors can be severe [3]. On the other hand, recent created sophisticated algorithms, such as Deep Neural Networks (DNNs), are slowly having great success in visual inspection in the high-volume manufacturing industry. This is due to their ability to automatically extract features within datasets unlike previous neural network algorithms which require handcrafting of labels before training [7]. A DNN algorithm of particular interest for automated visual inspection is the Convolutional Neural Networks (CNNs) due to their exceptional ability to extract features within images as well as learn relationships between these features to recognise entities of interest within an image. This has led to CNNs being the dominant algorithm in image processing and classification applications [7][8].

The use of CNNs for visual inspection and defects detection in aircraft wings has been carried out in literation [9]. However, the primary issue highlighted is the scarcity of data required to train the CNN. This problem in deep learning projects is even more pronounced in the high-value manufacturing industry, due to the relatively low quantity of products from which data could be generated.

### *2.2 Siamese Neural Networks (SNN)*

In many applications of CNNs, the aim is to classify an image into one of several classes [10]. However as mentioned earlier, the primary limitation of CNNs and DNNs in general are that they require very large quantities of data to train them [10] [11]. An alternative to the CNN architecture is the Siamese Neural Network (SNN) [12]. Rather than classifying images directly as in CNNs, SNN works by determining if the contents of two images are the same or not. SNNs have had a great deal of success in the fields of character recognition, face verification, and signature verification. They have performed very well when presented with a few or even a single training sample per class [12][13][14]. This performance is due to the nature of what SNNs learn from datasets. In standard CNNs, the network aims to learn the unique features that identify the class of an image [11], whereas in SNNs the network aims to learn the more basic features that describe an image and compare these features across pairs of images [12][13][14]. These features require much less data to train from and can also make use of transfer learning from other domains to start with some already learnt basic features, such as edges and corners [15]. The architecture of a SNN is made up of two identical input neural networks. The learnt features of these networks are extracted using standard convolutional layers as in a CNN. SNN achieves the task of image verification by comparing two images typically sampled randomly from the available classes of the dataset. The distance between extracted feature vectors in the two images are compared, and then the likelihood of them being of the same entity is produced at the output layer (See Figure 6).

### *2.3 Image Data Augmentation*

Although the volume of data in this industrial application area is relatively small, data augmentation is a useful tool that can be used to increase the size of the data set artificially. Data augmentation is usually applied to image data for two reasons. Firstly, to increase the size of the data set to prevent underfitting, and secondly, to introduce variability to the data set to prevent overfitting [16][17]. As supported by the state-of-the-art applications of SNNs for image verification, we focused on the use of basic image manipulations of geometric affine transformations, and colour space transformations [12] [13]. Affine transformations are geometric transforms applied to an image that can include flipping, cropping, translating, shearing, and rotating the image [12][17]. These transforms are sufficiently small and maintain important relationships between features within an image, ensuring that the image remains similar enough to be in the same class [17]. This is particularly important in this research where the aim is to verify the similarity of an input image to a reference image. However, the affine transform of cropping would not be suitable. This is because there is the potential for key areas of the image that characterise the correct installation of the component, to be cropped out. If a defect in the installation lies within this area of the image, the algorithm would have no way of identifying it and would consequently provide a false result. Finally, the alteration of the colour channels in an image allows the model to generalise better to images in different lighting conditions and further increases the size of the data set.

### *2.4 Applying pre-trained Neural Network via Transfer Learning*

A popular learning framework used in deep learning is transfer learning. The convolutional layers within a CNN learn increasingly complex features within the training images, starting with fundamental descriptors, such as edges and corners at the shallow layers, up to domain specific features near the output layer. Low level features such as edges and corners is often present in almost any image from any domain. Furthermore, mid-level features such as textures will be similar amongst images that are closely related.

Transfer learning involves taking a neural network pre-trained on a specific dataset and adapting it to a new dataset, potentially in a different application domain [15]. Transfer learning is also commonly employed in tasks where there is insufficient labelled training data to train a model from random initialisation, or where the distribution of the test data changes from that which it was trained upon [18]. It has been proven that transfer learning from a model trained on large-scale datasets, such as the ImageNet dataset, can yield much better results than training from scratch. Hence, this learning approach has become common practice for many classification tasks [19]. One such pre-trained model is the VGG16 architecture [20], which achieved state-of-the-art results on the ImageNet classification challenge 2014 and is now used commonly in transfer learning problems, where training data is highly limited and there is no existing large-scale dataset available. As SNNs consist of twin CNNs, transfer learning can be used for each network in the twin to utilize the feature extracting capability of the pre-trained model. These features can then be passed to the common comparison layers which can be trained specifically for the verification task.

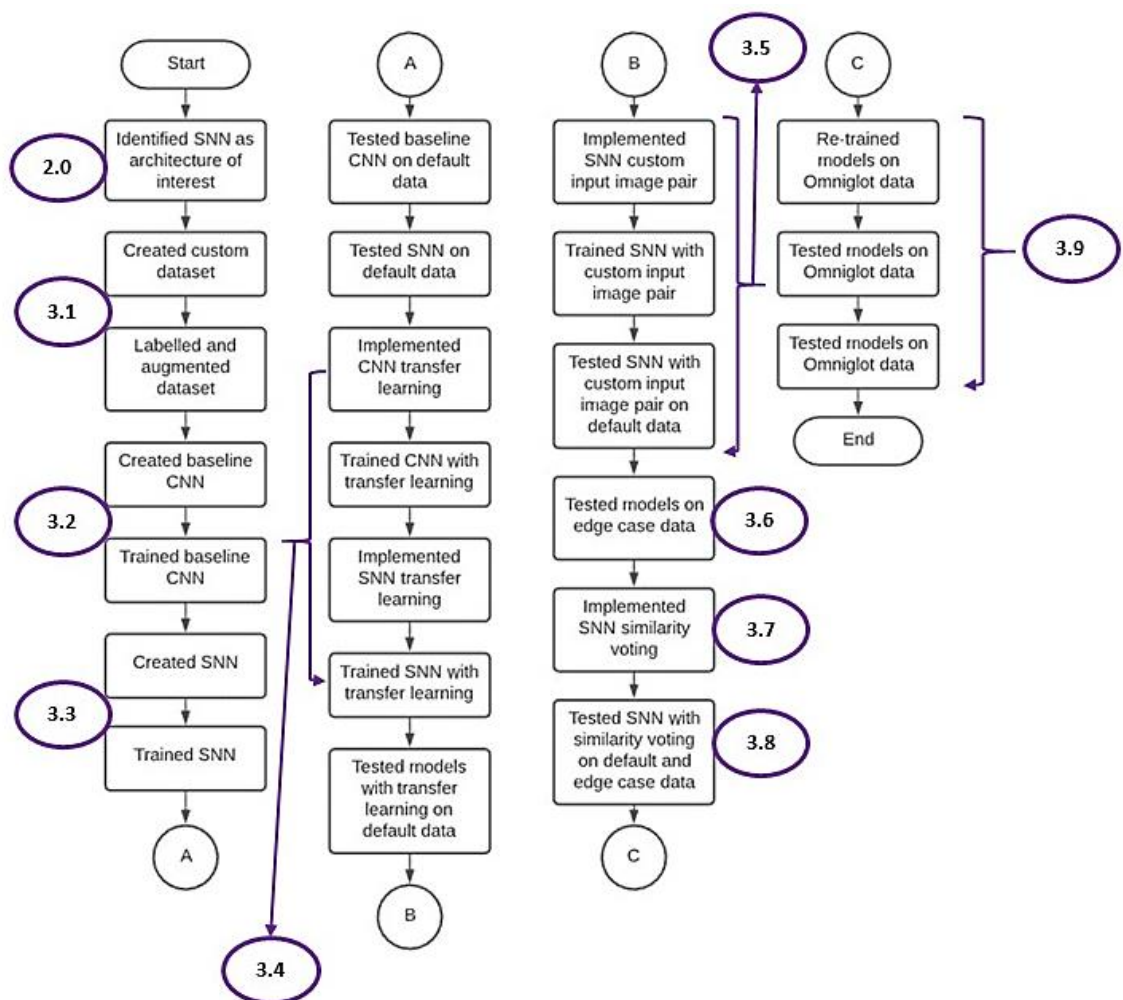

**Figure 3.** Overview of the process taken in our methodology. The numbers beside the boxes refer to the sections in the manuscript.

According to our knowledge, there lacks research into ensemble methods for SNNs specifically using bagged model voting. In this work, we apply SNN to the task of automatic verification of manually installed brackets while making use of limited amount of data to train the SNNs. We also investigate if ensemble voting could improve the performance of SNNs when testing for the similarity between a reference and test image. Towards this, we conduct a systematic investigation that compares our approach with a state of the art CNN architecture in order to ascertain if using SNNs have better performance for the automatic verification of manually installed brackets. According to our knowledge, this is the first time that SNNs have been used in this way.

## 3.0 Methodology

This section presents how experimental data was generated as well as how CNN and SNN models were constructed, trained and transfer learning applied to the models. We explain the process taken, tests conducted to assess the performance of the models developed as well as the technique applied to validate the results of our approach. The overview of our methodology is shown in Figure 3. It can be seen in Figure 3 that we created a series of models as follows: **(i)** a baseline CNN with VGG16 architecture; **(ii)** a novel SNN architecture that combines two of the baseline CNNs; **(iii)** a baseline CNN with transfer learning used; **(iv)** SNN architecture with transfer learning and **(v)** a SNN with similarity voting. In section 4, we compare the results of these models to ascertain which would be most suited for deployment in the automatic verification of installed brackets.

### *3.1 Component Representation and Data Generation*

A decision was taken to generate a custom dataset for this project because of the lack of open source databases that use brackets during aircraft installation. As a result, a suitable mock setup of the bracket installation process was developed. The replica bracket was created using LEGO® to represent the bracket and Play-Doh® modelling compound to represent the adhesive applied to the bracket's base. The Play-Doh® modelling compound was chosen over a true adhesive in order to ensure ease of removal and reapplication during experiments. A solid blank background with a slight shine and texture (created using a nylon sheet) was used to represent the wing surface. Furthermore, a suitable boundary marking was applied in order replicate the placement of the bracket process (see Figures 2 and 4).

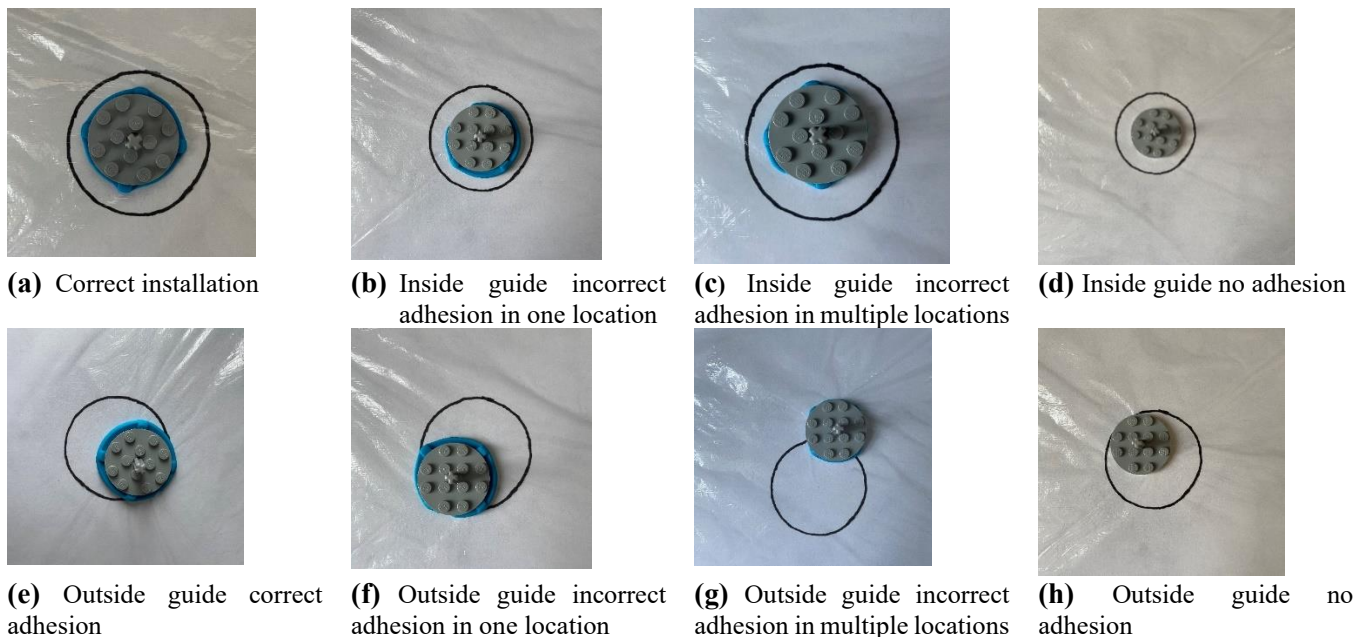

**(a)** Correct installation  **(b)** Inside guide incorrect adhesion in one location  **(c)** Inside guide incorrect adhesion in multiple locations  **(d)** Inside guide no adhesion

**(e)** Outside guide correct adhesion  **(f)** Outside guide incorrect adhesion in one location  **(g)** Outside guide incorrect adhesion in multiple locations  **(h)** Outside guide no adhesion

**Figure 4.** Example images of installation cases.

Experimental data was generated by taking images of the bracket installed both correctly and incorrectly. Since there will be variations in the way humans apply the brackets in the real world, consideration was given to the subtleties in each of the correct and incorrect installation cases. As a result, it was necessary to identify the sub-cases that might exist within both cases, and to ensure that data was gathered on as many variations as possible. This would support the generalization of the trained models. For the correct installation case, this involved all those installations which met the criteria as discussed previously (Figure 2).

Nevertheless, special attention was also paid to the edge cases, where the bracket was just within or outside the tolerance range from the marking. It was expected that these would be the most difficult for the models to distinguish between correct and incorrect installation cases. In the incorrect installation case, there were several sub-cases identified that the model would need to learn as belonging to the incorrect class (Figure 4). Images were obtained for each of these sub-cases, including images of the edge cases. By ensuring that all these cases were covered in the data generation, the chances of inherent bias in the dataset was mitigated. The generated dataset was validated and confirmed by an aerospace expert as what would be expected on a true production line. The images of the bracket dataset were RGB of shape (256, 256, 3). For training, there were 70 images of correctly installed brackets and 70 images of incorrectly installed brackets, totaling 140 training images. In order to replicate a low data scenario, 70 images of incorrectly installed brackets corresponding to 10 images of each of the 7 sub-classes identified previously were collected.

For validation, there were 100 images of correctly installed brackets and 100 images of incorrectly installed brackets, totaling 200 validation images. Furthermore, a framework to procedurally load the generated images and perform random augmentations was created. Random seeds were used to achieve variations in affine transforms of translation, rotation, scaling, shear, horizontal and vertical flipping as well as the colour space transform of scaling image brightness. Example augmentations of an image from the dataset can be seen in Figure 5, where the augmentations provide suitable variation in the images without altering the class properties. The images were also normalised to aid the gradient descent algorithm in minimising the loss function.

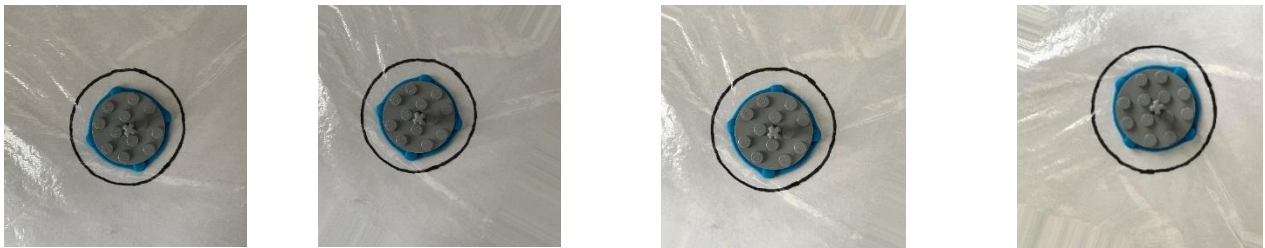

**Figure 5.** Original image of a correctly installed bracket (left). Three augmented images of the original image (right) generated by using random seeds in the rotation range = ±40°, height and width translation range = ±10%, Zoom range = ±20%, Shear range = ±20%, Brightness range = [70%, 130%]

### *3.2 Training a Baseline CNN Model from scratch*

A baseline CNN architecture was implemented (Figure 6a) and trained as a binary classifier. To each convolutional layer, we applied batch normalisation, a ReLU activation function, 30% dropout, and a Keras max-pooling layer with default settings. Optimisation of the model was achieved using the ADAM algorithm with a learning rate of 0.0001 and a binary cross-entropy loss function. Training was performed over 25 epochs with a batch size of 16. Images of either correctly installed brackets or incorrectly installed brackets were passed to the network during training, along with their class labels. The model predicted the probability of an image belonging to a correctly installed bracket class. A default threshold of 0.5 was applied to determine the binary class prediction. If above 0.5, the image was assumed to belong a correctly installed class and if below 0.5, the image was assumed to belong an incorrectly installed class.

### *3.3 Creating and training a SNN Model*

The SNN architecture was implemented with two baseline CNNs as seen in Figure 6b. A 30% dropout was applied to each of the CNN twin layers in the SNN. The twin layers were followed each by dense feature layer to which 30% dropout was also applied. This helped prevent overfitting during training. Optimisation of the SNN model was achieve using the same parameters as the baseline CNN models. Overall, the SNN was trained by passing pairs of images to the twin CNNs, along with the label of whether they were of the same class or not. The identical CNN networks with shared weights would then perform the same feature extraction on the images, producing feature vectors to be compared in the L1 distance layer. There were no learnable parameters in the L1 distance layer. The distance between the feature vectors were calculated as in Equation 1, where $p$ and $q$ are the two n-dimensional feature vectors. This would then be converted to a similarity score at the output layer by passing it through a sigmoid function, where a value closer to 1 denoted a higher confidence that the images were of the same class. A default threshold of 0.5 was used to determine whether the final decision was that the images were of the same class or not.

$$d_1(p, q) = \|p - q\|_1 = \sum_{i=1}^{n} |p_i - q_i| \qquad (1)$$

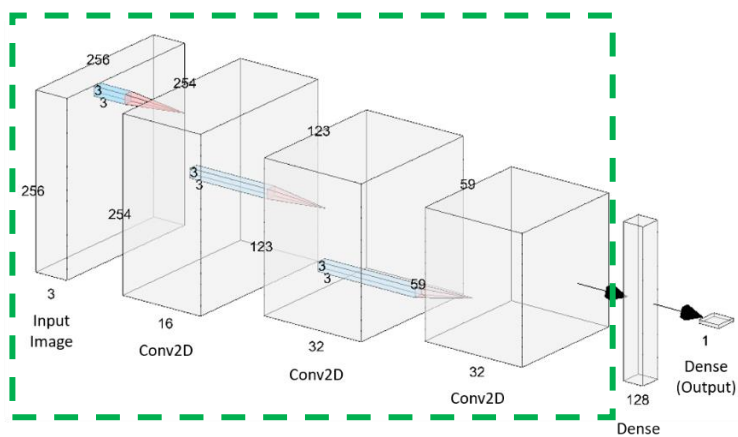

(a) Baseline Convolutional Neural Network Architecture based on the VGG16 model.

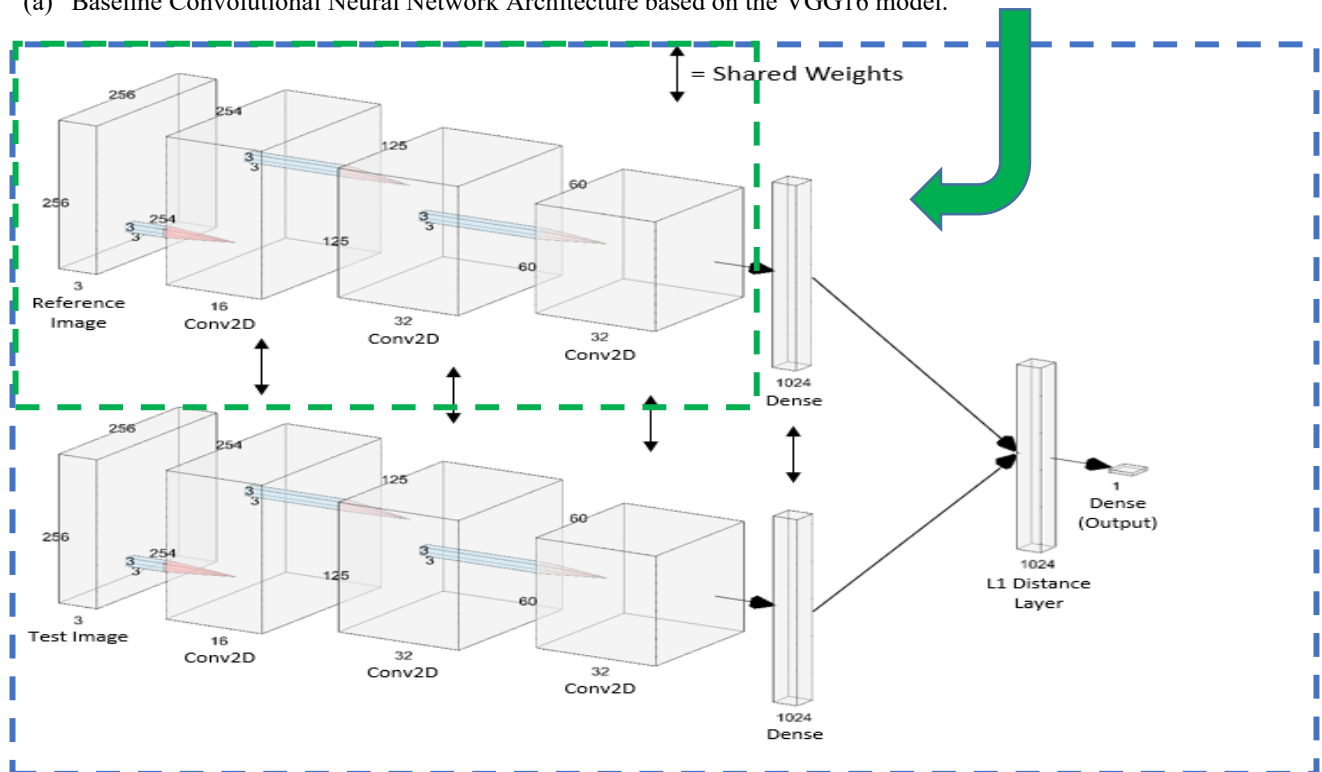

(b) Siamese Neural Network Architecture

**Figure 6.** Baseline CNN (Green box) and applying two CNNs (Blue box) to make up the SNN architecture.

### 3.4 CNN Model Training with Transfer Learning from a VGG16 model

The VGG16 model was selected for use as the baseline CNN in this study. This was due to its proven ability to work well during transfer learning. The VGG16 architecture is simple when compared to other state-of-the-art CNNs, such as ResNet [22] and Inception [23]. This enabled us to focus on evaluating the overarching SNN architecture instead of dealing with various aspects of the network. It also reduced the computational overheads required for training. The input layer of the VGG16 model was replaced with a new input layer suitable for taking the images in the bracket installation dataset. For investigating transfer learning, the earlier layers of the VGG16 model were set to remain fixed throughout training with the last block of convolutional layers set to be trainable.

Since it is known that earlier layers of deep neural networks mostly learn low-level features comprising of edges and corner, the transfer learning approach would enable specific high-level features of the brackets and adhesive to be captured and learned at the later layers. A randomly initialised dense layer of 128 units was added to the end of the VGG16 model to create the feature vector. The dense layer after the VGG16 model had 128 units with a dropout of 50%, and the model was optimised using ADAM with a learning rate of 0.0001 and a binary cross-entropy loss function. The model was trained over 25 epochs with a batch size of 16. For the SNN with transfer learning, each of the twin CNNs were replaced with the VGG16 model containing the transfer learnt parameters. The outputs of each VGG16 model were then passed to separate dense layers of 1024 units with 30% dropout to create the feature vectors. The remainder of the model then stayed the same as for the base SNN (without transfer learning). The model was optimised using the same optimiser and training options as before.

### 3.5 SNN Input Pair Consideration

The two images selected at each training step for an SNN are typically randomly sampled from the available classes. In this study, the image pairs could be one of: (correctly installed, correctly installed), (correctly installed, incorrectly installed), (incorrectly installed, correctly installed), or (incorrectly installed, incorrectly installed). Initially, this default method of selecting the images was used.

However, following on from our experimental results, the method of passing in the data was revisited. As discussed previously, the incorrectly installed bracket case had several sub-cases where each sub-case could differ significantly from the others. It was therefore hypothesized that attempting to label images from different incorrect sub-cases as being similar could cause optimisation problems for the SNN. An additional test was therefore conducted, looking to investigate the performance impact of the input image pair choice. For this, one of the input images was always presented as a reference image of a correctly installed bracket, and always to the same input of the network. The other input image could be of either a correctly installed bracket or any case of incorrectly installed bracket. This method of passing the input images is referred to as the "custom input image pair" from here on. The method was justified by the nature in which the SNN would be used in deployment. One image would be a reference image of a correctly installed bracket and the second would be of a newly installed bracket. The model would then verify whether the bracket had been installed correctly by comparing the images.

### 3.6 Discrimination of Edge Cases

As highlighted in Section 3.1, the correctly installed edge case images were of particular interest in this study. It was expected that these would be where most misclassifications of images by the models would occur and so specific tests were conducted to inform whether it is possible to identify the edge cases accurately. Of particular interest of the edge cases were those where the bracket is placed just within or outside the tolerance of the marked boundary and with correct adhesion (See Figure 7). As a result, additional images were taken and added to create a suitably sized dataset specifically for this edge case. The new edge case dataset had 50 images of incorrectly installed brackets for training and there were 40 images for validation.

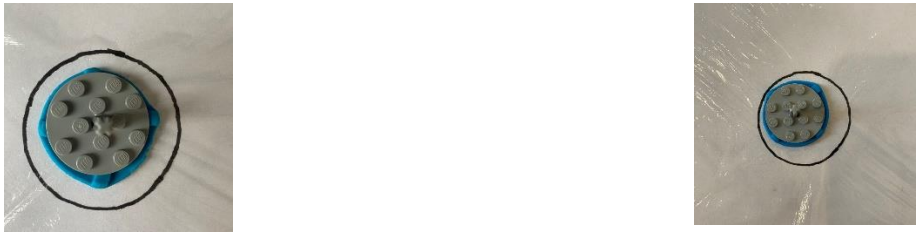

**Figure 7.** Example of a correctly installed bracket edge case (left). Example of an incorrectly installed bracket edge case (right)

### *3.7 SNN Similarity Voting*

When using the SNN to verify if a new test image is the same as a known reference image, the similarity score will depend on the reference image used. This means that the resulting decision of whether the images are the same or not could change depending on which reference image is used and how similar it is to the test image provided. Similarly, in deployment into aircraft assembly lines, verification of a bracket installation will be achieved by providing a reference image to the SNN of a correctly installed bracket, and then passing it an image taken of the newly installed bracket.

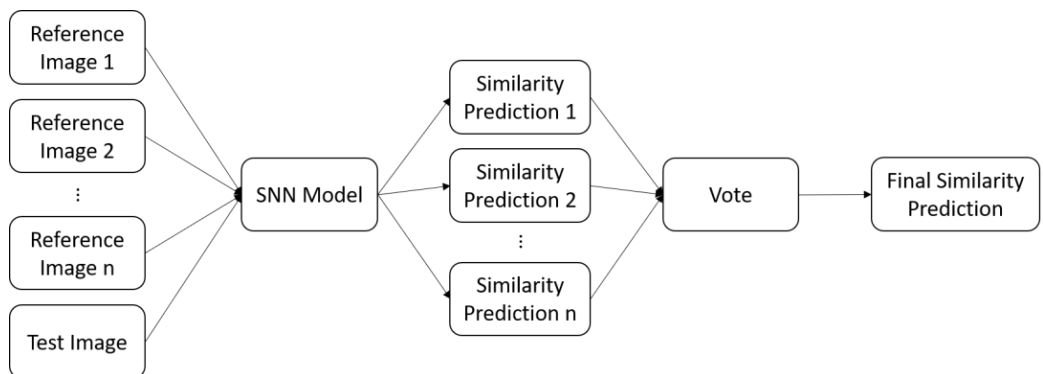

**Figure 8.** SNN similarity voting on reference images

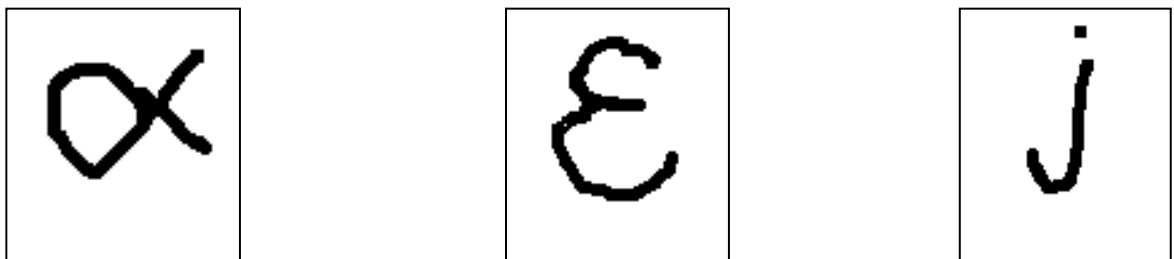

**Figure 9.** Example similar and dissimilar images in the custom Omniglot dataset. Reference image from the Greek alphabet (top). Similar image of a character from the Greek alphabet (bottom left). Dissimilar image of a character from the Latin alphabet (bottom right).

It is expected that following training, if a reference image of a correctly installed bracket and a similar test image were passed through a SNN, then both images would be identified as being similar, and the newly installed bracket (test image) would be verified as having been installed correctly. If however a different correctly installed bracket image were used as the reference image (e.g an image that is close to an edge case but not actually an edge case), the similarity score generated may be more uncertain and lie around the threshold value of 0.5. The bracket may then be classified as being installed incorrectly. In order to reduce the dependency of the results of the SNN model on the reference image used, we decided to use an ensemble approach in which a limited distribution of possible reference images were used as an input to the SNN model (Figure 8). The hypothesis here is that the dependency of the verification result on the reference image used would be reduced if the test image were compared to multiple reference images. A majority voting rule would then produce the correct result. Compared to the typical bagging method in which predictions from multiple machine learning algorithms come together to make more accurate predictions than any individual model, in our work, multiple reference-test input pairs are evaluated on a single SNN model and a vote on the final outcome is used for verification of the bracket installation process.

### *3.8 Testing and Evaluation*

In order to evaluate and compare the various models developed (See the process diagram in Figure 3), suitable metrics were used to assess the models. Though the SNN performed image verification rather than classification, the typical classification performance metrics still applied. In particular, the accuracy (Equation 2), precision (Equation 3), and recall (Equation 4) metrics were selected. The accuracy metric was selected as an initial indicator of model performance, which allowed for the training progress to be monitored and simple comparisons to be made between models. However, accuracy by itself is not a sufficient indicator of performance. As a result, the precision and recall metrics were also observed.

The chosen metrics enabled quantitative assessment of models performance and also revealed important industrial implications. The precision of any model would indicate what proportion of brackets identified as being incorrectly installed were actually incorrectly installed. From another perspective, it could be inferred from this value how often a bracket identified as being incorrectly installed was in fact correctly installed. This metric, known as the False Discovery Rate (FDR) (Equation 5), is calculated from the precision metric. Aerospace companies would be interested in both the precision and FDR metrics because they would indicate how often a human operator would have to give a second opinion on the automated test. If the precision was too low, or conversely the FDR was too high, then the solution would be returning many false positives (FPs). The use of trained models in the verification process would then not be cost effective, as human operators would still be a common necessity in the procedure. Perhaps the most important metric in safety critical sectors like Aerospace is the recall. Practically, this metric shows what proportion of incorrectly installed brackets are identified as such. From a different perspective, the number of incorrectly installed brackets falsely identified as being correctly installed can be evaluated by another metric, the False Negative Rate (FNR). A low recall or high FNR value would indicate many incorrectly installed brackets were passing through undetected, which could have severe consequences should aircraft be put through to operation with these onboard.

$$Accuracy = \frac{TP}{TP + FP + TN + FN} \qquad 2$$

$$Precision = \frac{TP}{TP + FP} = Positive\ Predicted\ Value\ (PPV) \qquad 3$$

$$Recall = \frac{TP}{TP + FN} = Sensitivity = True\ Positive\ Rate\ (TPR) \qquad 4$$

$$FDR = \frac{FP}{FP + TP} = 1 - Precision \qquad 5$$

$$FNR = \frac{FN}{FN + TP} = 1 - Recall \qquad 6$$

### *3.9 Results Verification on a Comparable Dataset:*

In order to give confidence in the results obtained using the custom bracket dataset generated, the decision was made to repeat the experiments on a well-known dataset and use the results as a baseline. It could then be seen whether the same trends were present in the results. This was an important additional test for this study, where the data used was created specifically for the bracket installation task. It would also prove that the previous results obtained from the generated bracket dataset were not influenced by bias. Towards this, the Omniglot dataset was selected. This dataset is a popular benchmark for SNNs [12]. To make the dataset similar to that used in the bracket verification task, two alphabets that were similar in nature to each other were chosen. The Latin and Greek alphabets were chosen and this enabled us to maintain the similarity and dissimilarity that was present between correctly and incorrectly installed brackets. Examples of similar and dissimilar images for this task are seen in Figure 9, where in this case similarity refers to whether the character is from the same alphabet or not. With this new dataset, the task was to perform alphabet verification at the alphabet level because each alphabet set was analogous to correctly or incorrectly installed bracket classes, while each character in an Alphabet set was analogous to the different sub-classes of brackets as described previously.

## 4.0 Results, Analysis and Discussion

This section presents results obtained, comparisons between CNN and SNN, analysis on the effectiveness of transfer learning, similarity voting and what the results mean in an industrial context. The Python programming language was used along with Keras API and a TensorFlow backend in this work. This enabled access and customization flexibility on a vast array of matured deep learning tools and model structures.

Table 1. Performance of the baseline CNN

|  | Baseline CNN with Random Initialisation | | Baseline CNN with Transfer Learning | | Edge Case Performance of Baseline CNN | |
|---|---|---|---|---|---|---|
|  | Training | Validation | Training | Validation | Training | Validation |
| **Accuracy** | 88% | 50% | 98% | 75% | 96% | 85% |
| **Precision** | 87% | 50% | 96% | 97% | 98% | 77% |
| **Recall** | 86% | 100% | 96% | 54% | 94% | 99% |
| **FDR** | 13% | 50% | 4% | 3% | 2% | 23% |
| **FNR** | 14% | 0% | 4% | 46% | 6% | 1% |

### *4.1 Comparison of Models Trained from Random Initialisation*

The validation metrics of the baseline CNN trained from random initialisation showed that the model had learned to classify every validation image as being of an incorrectly installed bracket (Table 1). This is evident from the 100% recall and 50% precision, meaning that all images of incorrectly installed brackets were identified, but only 50% of the images identified as being of incorrectly installed brackets actually were. When compared to the validation performance of the SNN trained from random initialization (See Table 2), the CNN's accuracy indicated that the performance was no better. Hence, SNN was no better but no worse than the baseline CNN. Also, it became apparent that although the SNN did not require many images per class to train, it still required a considerable number of images in total in the dataset if it were to be trained from random initialisation. Consequently, we focused on approaches using transfer learning to allow for more meaningful results to be obtained and for better comparisons of the architectures.

### *4.2 Comparison of Models using Transfer Learning*

As informed by [15], it was expected that transfer learning would boost performance of both the CNN and SNN, particularly in this low-quantity data task. Firstly, when comparing the CNN with transfer learning to that trained from random initialisation, there was an increase in overall validation performance as shown by the improvement in accuracy in Table 1. Although the recall reduced, this could be due to the classifier attempting to correctly classify the correctly installed brackets. Next, comparing the SNN with transfer learning to the SNN model without transfer learning (Table 2), it was discovered that there was a slight performance increase observed. This was seen in the increase in accuracy as well as increase in the precision.

Nevertheless, there was a decrease in the recall. This was a concerning result for the SNN, potentially indicating that it would not be appropriate for the verification task.

Table 2. Performance of the SNN

|  | SNN with Random Initialisation | | SNN with Transfer Learning from VGG16 model | | SNN with Custom Input Image Pairs and TL | | SNN with Similarity Voting and TL from VGG16 model | | Edge Case Performance of SNN with TL | | SNN on the edge case using similarity voting, CIIP, & TL from VGG16 model | |
|---|---|---|---|---|---|---|---|---|---|---|---|---|
|  | Training | Validation | Training | Validation | Training | Validation | Training | Validation | Training | Validation | Training | Validation |
| **Accuracy** | 50% | 49% | 56% | 57% | 90% | 76% | N/A | 81% | 90% | 67% | N/A | 76% |
| **Precision** | 50% | 51% | 62% | 70% | 90% | 77% | N/A | 84% | 89% | 80% | N/A | 94% |
| **Recall** | 47% | 79% | 44% | 16% | 91% | 77% | N/A | 95% | 94% | 47% | N/A | 80% |
| **FDR** | 50% | 49% | 38% | 30% | 10% | 23% | N/A | 16% | 11% | 20% | N/A | 6% |
| **FNR** | 53% | 21% | 56% | 84% | 9% | 23% | N/A | 5% | 6% | 53% | N/A | 20% |

However, after the consideration was given to the operation of the SNN as discussed in Section 3.5, the SNN performance was improved. Table 2 summarises the comparisons between models trained with transfer learning. Comparing the two architectures (CNN and SNN) with transfer learning revealed that CNN outperformed the SNN across all metrics. Nevertheless, SNN performance improved when custom input image pairs where used versus when images pairs were not used. By custom input image pair, we mean having one input of the SNN have a reference image of a correctly installed bracket with the other input of the SNN receiving both correctly installed and incorrectly installed brackets.

Table 3. CNN and SNN performance on omniglot dataset

|  | Baseline CNN on Omniglot Dataset | | SNN with Custom Input Image Pairs and TL | | SNN on the edge case using similarity voting, CIIP, & TL from VGG16 model | |
|---|---|---|---|---|---|---|
|  | Training | Validation | Training | Validation | Training | Validation |
| **Accuracy** | 89% | 82% | 80% | 74% | N/A | 82% |
| **Precision** | 90% | 83% | 76% | 85% | N/A | 93% |
| **Recall** | 87% | 79% | 82% | 63% | N/A | 88% |
| **FDR** | 10% | 17% | 24% | 15% | N/A | 7% |
| **FNR** | 13% | 21% | 18% | 37% | N/A | 12% |

### *4.3 Effects of Input Image Pairs on SNN Performance*

With the use of custom input image pairs, SNN was seen to perform equally as well as the CNN in validation. The CNN was seen to be more precise than the SNN with near-perfect precision, however the SNN achieved higher recall (Table 2). These results did not prove either way whether the SNN was better or worse than the CNN at performing visual verification with low quantities of data. However, from an industrial perspective the differences in precision and recall have significant implications.

Although the SNN did not prove better or worse than the CNN for this task, the significant improvement in SNN performance when using the custom input image pair highlighted the importance of considering how the data is presented to a SNN. The improvements seen show that the variations in the incorrect installation sub-cases were the source of the model instability during training. This was supported further by analysing the training graphs for the SNN with and without the new method of passing the input images. Figure 10 shows the instability during training and that no real convergence was reached when custom input image pairs were not used. Figure 11 on the other hand shows a smooth training process and that convergence was reached when custom input image pairs were used. In Figure 10, the variations in the incorrectly installed class and its sub-classes likely hindered the CNN performance. The solution to this problem (namely the use of a custom input image pair) was one that is specific to the SNN architecture and CNNs would not be able to utilise this solution because of their single input.

### *4.4 Effectiveness of Similarity Voting in SNNs*

The novel similarity voting method achieved significant improvements in SNN validation performance (Table 2) with improvements across all metrics especially in recall. An increase in recall is often accompanied by a reduction in precision and vice versa (known as the precision-recall trade-off) [24], however, here it was seen that the similarity voting method could improve both metrics in conjunction. This naturally led to an increase in the accuracy metric as well. This is because of the way in which the voting system handles new test images. When provided with a new test image, the model would have more confidence in labelling it as an incorrectly installed bracket if it were dissimilar to the majority of reference images of correctly installed brackets. This reduced the number of FPs due to the rigor in the decision to label a bracket as being incorrectly installed. As the precision is inversely dependent on the number of FPs, this led to the increase in precision. The same logic applies to the FN cases. The model would less likely misclassify an incorrectly installed bracket as correctly installed because it is unlikely that it would be similar to a majority of the reference images. The number of FNs would therefore decrease leading to an increase in recall. Since the biggest increase in performance came from the increase in recall, this was likely the primary source of error for the SNN before using similarity voting.

Comparing the best performing SNN with similarity voting to the best performing CNN with transfer learning, it was seen that the SNN now outperformed the CNN in general accuracy. It still did not match the very high precision of the CNN, but vastly outperformed it in recall. With respect to the initial hypothesis of the research, these results showed that with additional methods applied, the SNN could outperform the CNN in the task of visual inspection of installed components.

Similarity voting was one of several additional methods that could be applied to the SNN to increase performance. The SNN architecture could potentially be considered when developing and implementing visual verification algorithms, particularly in low-quantity data scenarios.

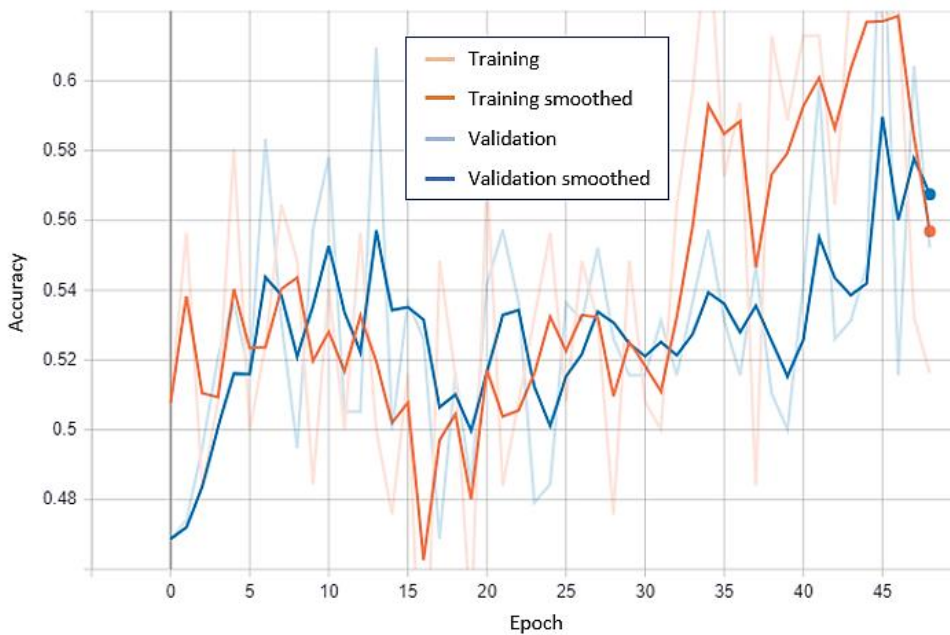

**Figure 10.** Accuracy curve for SNN with transfer learning with unstable training and no convergence.

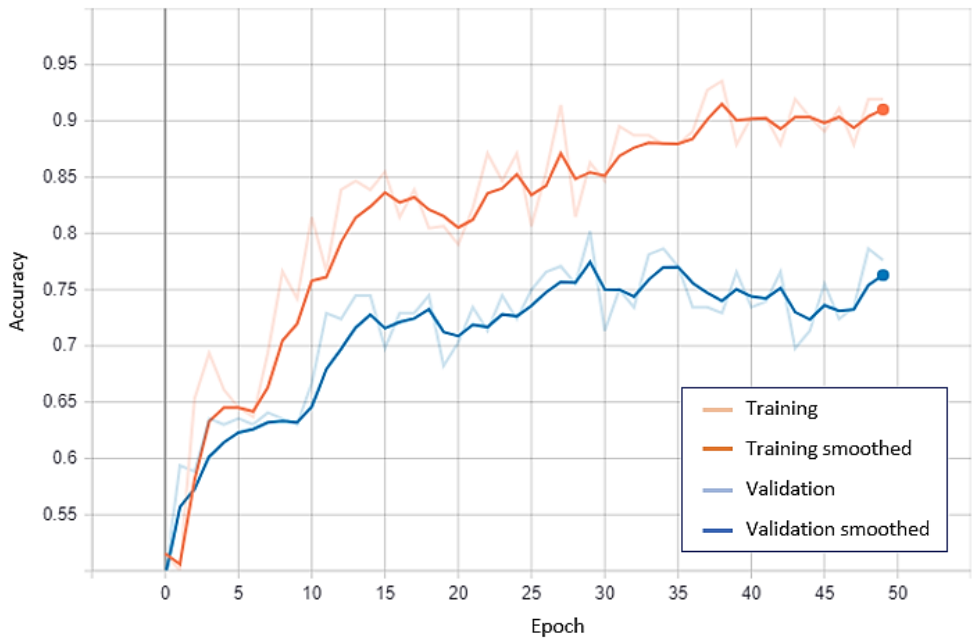

**Figure 11.** Accuracy curve for SNN with transfer learning and custom input image pair, showing stable training and convergence.

### *4.5 Edge Case Discrimination*

The results obtained by both the CNN and SNN on the edge case data confirmed that it is possible for these models to discriminate between the edge cases, that is, brackets just within and just outside the tolerance. The CNN with transfer learning performed well as seen in Table 1. This was likely due to there being less variation in the incorrectly installed case, as the only incorrect sub-case included in this set was for brackets with correct adhesion but placed just outside the tolerance. This higher correlation between the images in that class would have then led to more stable training of the network, yielding the better performance seen. The SNN with transfer learning and the custom input image pair underperformed. However, with similarity voting this drastically improved.

### *4.6 Observations from using Omniglot Dataset for Verification*

Repeating the primary experiments on the custom Omniglot dataset yielded reassuring results. Starting with the CNN and SNN, both with transfer learning and the SNN with the custom image input pair, they were both seen to perform similarly to how they did on the bracket installation dataset. The CNN did perform better than previously, but not to the extent that would indicate the CNNs superiority for the task. When the similarity voting was introduced for the SNN, a similar result to that seen on the bracket installation dataset was observed as seen in Table 3. Again, it was recall that received a significantly large increase, and this led to the conclusion that the similarity voting method introduced in this study could have significant potential.

## 5.0   Applying results in an industrial context

One of the goals of this work was to investigate the appropriate methods for performing visual inspection of installed components when little data was available for model training. When using transfer learning, the validation precision achieved was high, which in an industrial context would mean few FPs would need investigating unnecessarily by human operators, making the solution more cost-effective. However, the recall achieved was low and would not be acceptable industry, as this would imply that nearly half of all incorrectly installed brackets were passing through inspection unnoticed. There would however be opportunities to improve the recall if further methods were employed, such as ensemble models. As such, the use of CNNs should not be ruled out until further research. SNN, particularly with similarity voting or transfer learning, is potentially a strong candidate in low-quantity data scenarios. It achieved very high recall, which is an important metric for the safety regulators of aircraft manufacturing companies. This would give confidence that the number of incorrectly installed brackets making it onto operational aircraft unnoticed would be low, especially considering that the frequency of incorrectly installing brackets is low as it is. The precision achieved by the SNN was relatively high, although perhaps not to the level that would make the solution usable in industry. Additional work should attempt to further improve precision, and possibly recall using various techniques.

## 6.0 Conclusions and Future Work

The contributions made in this research apply to both academia and to industry. From an academic perspective, we investigated the effectiveness of SNN for image verification with low quantities of training data. It was shown that the SNN architecture was suitable and could achieve a reasonable level of performance when compared with CNN and potentially better when additional methods are applied. It was confirmed that transfer learning can allow both architectures to significantly increase their level of performance, even with low quantities of data. From an industrial perspective, our contributions can be applied to the aircraft manufacturing industry when visual inspection processes on assembly lines need to be automated. Specifically, an image verification approach has been presented that could take in a test image of a newly installed component, whether installed by a human operator or autonomously, and verify whether it has been installed correctly. Considering the safety nature of the aircraft manufacturing industry, performance metrics were observed that could be attractive to the sector. According to our knowledge, this is the first time a detailed study of this sort has been carried out in the use of Deep Neural Networks to automatically verify manually installed brackets in the aerospace sector.

In the future, we would implement and use the contrastive loss function for training the SNNs, as opposed to the binary cross-entropy loss function used herein. This is due to claims in [37] that it significantly improves the performance of SNNs. Also, other alternative state-of-the-art CNNs, such as ResNet and Inception, will be trialed in place of the VGG16 model when transfer learning. From an industrial perspective there would be two more avenues of interest. The first of these would be to apply the methods of this study to images of the real brackets used on aircraft wings. Similar trends in the results would be expected, verifying the methods' applicability to the real-world problem. Furthermore, the results would prove that the proposed solution can be transferred to variations of the bracket in new aircraft models, without the need for retraining leading to cost savings.

## Acknowledgements

The authors will also like to acknowledge the Royal Academy of Engineering under the Research Chairs and Senior Research Fellowships scheme. Professor Ashutosh Tiwari is the Research Chair in Digitization for Manufacturing at the University of Sheffield.